\documentclass[11pt]{article}

\usepackage[final]{acl}

\usepackage{times}
\usepackage{latexsym}
\usepackage[T1]{fontenc}
\usepackage[utf8]{inputenc}
\usepackage{microtype}
\usepackage{inconsolata}
\usepackage{graphicx}
\usepackage{url}
\usepackage{amsmath}
\usepackage{amsthm}
\usepackage{booktabs}
\usepackage{tabularx}
\usepackage{xcolor}
\usepackage{ragged2e}

\providecommand{\hg}[1]{}
\providecommand{\psj}[1]{}
\providecommand{\dw}[1]{}
\providecommand{\Neeraj}[1]{}

\title{Code-Mixing and Code-Switching for Text in the LLM Era:\\
A Playbook of Models, Data, Evaluation, and Open Problems}

\author
{Himanshu Gupta $^{1}$ \quad Pratik Jayarao $^{2}$ \quad Neeraj Varshney $^{1}$ \quad \textbf{Chaitanya Dwivedi} $^{2}$  \\
\small{$^{1}$Arizona State University} \quad
\small{$^{2}$Carnegie Mellon University} \quad \\
\tt\small {\{hgupta35,nvarshn2\}}@asu.edu \\
\tt\small {\{pjayarao,cdwivedi\}}@alumni.cmu.edu
}

\begin{document}
\maketitle

\begin{abstract}
Code-mixing and code-switching (CSW) remain challenging phenomena for large language models (LLMs).
Despite recent advances in multilingual modeling, LLMs often struggle in mixed-language settings, exhibiting systematic degradation in grammaticality, factuality, and safety behavior.
This survey provides a comprehensive overview of CSW research in the era of LLMs.
We introduce a unifying taxonomy that organizes prior work along dimensions of data, modeling, and evaluation, and we distill these findings into a practical playbook of actionable recommendations for building, adapting, and evaluating CSW-capable LLMs.
We review modeling approaches ranging from CSW-tailored pre-training and task-specific post-training to prompting strategies and in-context learning.
We analyze current evaluation practices, highlighting sources of instability and limited reproducibility, and we catalog existing benchmarks while critically examining their linguistic coverage and English-centric biases.
Finally, we discuss emerging safety concerns, including use of code-mixing as a mechanism for bypassing model safeguards, and identify open research challenges.
\footnote{The work does not relate to authors' position at Amazon}
\end{abstract}

\section{Introduction}

Code-mixing and code-switching (CSW) is the practice of alternating between two or more languages within a conversation or a sentence, and it is a ubiquitous linguistic phenomenon in multilingual communities. While CSW is the norm for millions of users on social media and messaging platforms, it remains a formidable challenge for Natural Language Processing (NLP) systems trained primarily on monolingual corpora.
The research landscape has shifted dramatically with the advent of Large Language Models (LLMs). Previous eras relied on fragmented, task-specific approaches, from statistical models (HMMs) to specialized neural architectures (RNNs) which required extensive feature engineering or fine-tuning for each language pair. The LLM era promises a unified solution: a single foundational model capable of handling diverse tasks. However, despite their scale, current LLMs frequently fail when confronted with the intricacies of mixed-language inputs. These failures manifest not only as performance degradation in standard tasks but also as ``language confusion,'' where models unexpectedly revert to a dominant language like English, and ``hallucination'' of cultural nuances \citep{zhang2023notyet, Mohamed2025LostITD}.
Furthermore, CSW has emerged as a critical vector for AI safety. Recent red-teaming studies reveal that safety guardrails aligned on English data are brittle; malicious actors can effectively ``jailbreak'' models using code-mixed prompts that bypass semantic filters \citep{Yoo2024CodeSwitchingRLR}. This creates an urgent need to understand how LLMs process mixed inputs and to develop robust adaptation strategies.

\begin{figure}[t!]
    \centering
    \includegraphics[width=\linewidth, height=4cm]{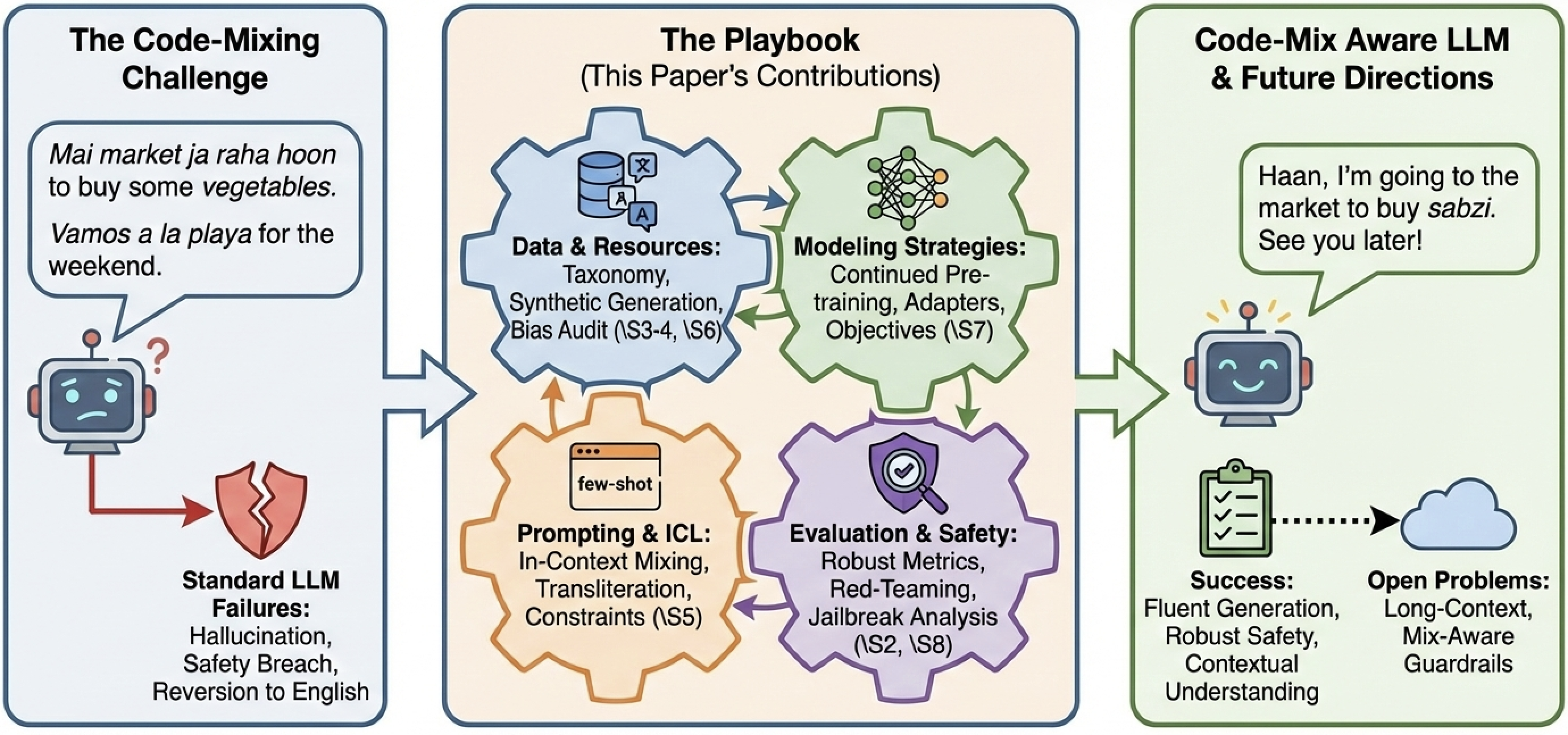}
    \caption{A comprehensive framework for addressing code-mixing in the LLM era, spanning data, modeling, prompting, and evaluation to build more robust and safe multilingual systems.}
    \label{fig1:teaser}
    \vspace{-0.5cm}
\end{figure}

This survey maps this evolving field through a methodological taxonomy. Unlike broader multilingual surveys \citep{sheth2025beyond}, we focus specifically on the LLM lifecycle. While recent comprehensive surveys provide broad taxonomies of the research landscape \citep{sheth2025beyond,winata2023decades,Doruz2023ASOU}, our work differentiates itself by structuring findings as an operational playbook centered on the LLM lifecycle. We move beyond a literature-centric review to offer actionable, practitioner-focused recommendations for data generation, prompting strategies, and safety alignment to effectively build and deploy code-mixed systems.
We begin by auditing \textbf{Data Resources} (\S\ref{sec:Background_Landscape}) for biases such as English centricity. We then analyze \textbf{Modeling Strategies} (\S\ref{sec:Prompting}--\S\ref{sec:Training}), contrasting the efficacy of prompting techniques against structural interventions like continued pre-training. Finally, we address the downstream implications: the instability of current \textbf{Evaluation} metrics (\S\ref{sec:Evaluation}) and the imperative of \textbf{Safety and Red-Teaming} (\S\ref{sec:Safety}).

\section{Background: Linguistic Foundations and Computational Landscape}
\label{sec:Background_Landscape}

In this section, we consolidate the core linguistic definitions, the typology of language interactions, and the broad landscape of NLP tasks associated with code-mixing.

\subsection{Core Definitions}
\label{subsec:Core_Definitions}

We first define the key linguistic concepts utilized in the study of mixed-language data:

\noindent\textbf{Code-Switching vs. Code-Mixing:} While often used interchangeably, \textit{Code-Switching} technically refers to inter-sentential alternation (between sentences), whereas \textit{Code-Mixing} denotes intra-sentential blending (within a clause).

\noindent\textbf{Matrix vs. Embedded Language:} Derived from the Matrix Language Frame (MLF) theory, the \textit{Matrix Language} provides the morphosyntactic frame (e.g., word order), while the \textit{Embedded Language} contributes lexical items.

\noindent\textbf{Transliteration:} The representation of text from one script in another (e.g., Hindi written in Latin script), a dominant feature of digital CSW. This introduces high orthographic variance, complicating subword tokenization.

\noindent\textbf{Grammatical Constraints:} Intra-sentential switching is not arbitrary. It adheres to rules such as the \textit{Equivalence Constraint} (switches occur where surface structures of both languages map onto each other) and the \textit{Free Morpheme Constraint} (switches are prohibited between a root and a bound morpheme) \citep{poplack1980typology}.

\subsection{A Typology of Code-Mixing}
\label{subsec:Typology}

The CSW literature spans a spectrum from bilingual to multilingual settings. The majority of work remains English-centric, covering South Asian pairs (Hinglish, Tanglish) in Romanized scripts, European pairs (Spanish-English), and East Asian pairs mixing English with logographic systems. Beyond bilingual mixing, recent studies explore tri-lingual environments (e.g., English-Bangla-Hindi), intra-family mixing (e.g., Spanish-Portuguese), dialectal switching (e.g., MSA-Egyptian Arabic), and ``hyper-mixed'' synthetic scenarios for stress-testing model safety.

\noindent CSW is not a single task but a pervasive condition affecting the entire NLP stack---from token-level language identification to generation and retrieval. A detailed landscape of affected tasks and applications is provided in Appendix~\ref{app:Tasks}.

\section{Prompting and In-Context Learning}
\label{sec:Prompting}
With the rise of large language models (LLMs), prompting and in-context learning (ICL) have become the first line of defense for tackling linguistic tasks. In context of code-mixing, literature reveals a dual utility for prompting. First, code-mixed and transliterated prompts serve as powerful ``cognitive bridges'' to unlock better multilingual performance in low-resource scenarios. Second, when goal is to generate code-mixed text itself, specialized constraints are required to prevent models from reverting to monolingual English. We synthesize these two distinct prompting paradigms below.

\subsection{Code-Mix Prompting to Unlock Multilingual Capabilities}
A growing body of research suggests that code-mixing is not just a target task but a mechanism to improve an LLM's general multilingual reasoning and transfer capabilities, particularly for low-resource languages.

\paragraph{Bridging Representations via In-Context Mixing.}
Research shows that mixing languages within few-shot exemplars aligns internal representations for low-resource languages. \citet{shankar-etal-2024-context} demonstrate \textbf{In-Context Mixing (ICM)}, where replacing English content words with target equivalents significantly improves downstream tasks like grammatical error correction by bridging the semantic gap.
Similarly, \citet{Zhu2025CSICL} propose \textbf{Code-Switching In-Context Learning (CSICL)}, using examples that progressively transition from the target language to English. This acts as a cognitive ramp, reducing the model's reliance on implicit internal translation and improving cross-lingual transfer.

\paragraph{Overcoming Script Barriers via Transliteration.}
For languages using non-Latin scripts (e.g., Hindi, Bengali, Arabic), LLM tokenizers often fragment words into meaningless sub-tokens. \citet{ma2024exploringroletransliterationincontext} identify \textbf{transliteration} as a potent strategy to overcome this. Converting non-Latin script examples into the Latin alphabet (Romanization) increases lexical overlap with the model's English-centric vocabulary. Providing Romanized text in the prompt---or both native and Romanized scripts simultaneously---maximizes the semantic signal for sequence labeling and classification tasks.

\paragraph{Activating Cultural Knowledge.}
Prompting strategy also dictates knowledge retrieval. \citet{Kim2024CanCTO} find that the language of the prompt serves as a context trigger. Posing a question in a code-mixed format can ``activate'' specific cultural neurons and localized knowledge that remain dormant during English-only prompting. Thus, code-mixed priming is essential when the downstream task involves cultural reasoning or regional sentiment.

\noindent\paragraph{Practitioner's Insight: The Multilingual Unlock}
\begin{itemize}
\item \textbf{If the model struggles with a low-resource language:} Do not use purely monolingual prompts. Inject target-language words into English few-shot examples (ICM) to bridge the semantic gap.
\item \textbf{If the language uses a Non-Latin script:} The tokenizer is likely your bottleneck. Transliterate the input to Latin script in the prompt to leverage the model's English-centric pre-training.
\item \textbf{If task requires cultural nuance:} Use code-mixed queries to ``wake up'' region-specific parameters and reduce generic westernized hallucinations.
\end{itemize}

\subsection{Prompting Strategies for Controlled Code-Mixed Generation}
When the objective shifts to generating code-mixed text (e.g., for chatbots or synthetic data), naive prompting often fails, leading to ``unnatural'' mixing or language reversion. Successful generation requires explicit constraints.

\paragraph{Explicit Definitions over Role-Playing.}
Vague instructions like ``act as a bilingual speaker'' are often insufficient for maintaining a mixed output. \citet{Zhang2023PromptingMLB} show that including a formal definition of code-mixing in the system prompt yields significantly higher quality output than simple role-playing prompts.

\paragraph{Linguistic and Syntactic Constraints.}
To prevent random, unnatural switching, prompts must enforce structural rules. \citet{Kuwanto2024LinguisticsTMG} demonstrate that injecting explicit linguistic constraints (e.g., instructing the model to only switch at specific syntactic boundaries like Noun Phrases) drastically improves human-perceived naturalness. This approach leverages the LLM's instruction-following capability to adhere to theoretical frameworks like the Equivalence Constraint.

\paragraph{Rule-Based Construction for Data Synthesis.}
For creating evaluation benchmarks or stress-testing systems, organic generation can be too unpredictable. \citet{Gupta2024CodeMixerYNL} suggest abandoning natural generation in favor of rigid, rule-based prompts (e.g., ``Substitute every noun with its English translation''). While stylistically rigid, this method guarantees the presence of code-mixing at specific densities, which is crucial for rigorous model evaluation.

\noindent\paragraph{Practitioner's Insight: The Prompting Playbook}
\begin{itemize}
\item \textbf{For Assistants and Natural Interaction:} Avoid vague persona prompts (``You are a bilingual speaker''), rather provide formal definitions in the prompt.
\item \textbf{To Prevent Unnatural Switching:} Explicitly enforce structural constraints in prompt (e.g., allow switches only at noun phrases or clause boundaries). This discourages token-level flickering and produces code-mixing that better matches linguistic patterns and human acceptability.
\end{itemize}

\paragraph{Limitations} Even optimized prompts often lag behind specialized fine-tuned models for structural tasks like Language Identification (LID) \citep{zhang2023notyet, Mohamed2025LostITD}. Therefore, prompting should be viewed as a high-utility prototyping tool, while production-grade systems often require architectural interventions discussed in \S\ref{sec:Training}.

\section{Strategies for Large-Scale Code-Mixed Data Generation}
\label{sec:Methods}

The scarcity of naturally occurring, labeled code-mixed data has motivated a wide spectrum of generation strategies. The most effective approaches balance three competing goals: (i) \emph{coverage} across language pairs and domains, (ii) \emph{quality} and naturalness of mixing patterns, and (iii) \emph{control} over where, how often, and why switching occurs.

\paragraph{Alignment-guided substitution (cheap scaling baseline).}
A compute-efficient baseline is \textbf{Alignment-Guided Substitution}, which uses word- or phrase-level alignments (e.g., GIZA++, FastAlign) to map spans across bilingual resources and swap selected tokens or phrases to induce switching.
To prevent rigid phrasing or invalid boundaries common in naive swaps, practitioners enforce lightweight constraints (e.g., POS/NER restrictions, aligned-phrase swapping) to preserve meaning and fluency.

\paragraph{Generalizable zero-shot synthesis with multilingual backbones.}
When a multilingual pre-trained model is available, \textbf{Generalizable Zero-Shot Synthesis} enables rapid scale-up to new language pairs without curating large code-mixed corpora per pair.
The \textit{GLOSS} framework demonstrates that freezing the backbone and training only a lightweight parameter-efficient module (e.g., adapters) can support synthesis that transfers to unseen language pairs \citep{hsu-etal-2023-code}.

\paragraph{Filter-then-finetune for high-fidelity scaling.}
A recurring failure mode in large synthetic corpora is ``hallucinated'' or unnatural mixing. The \textbf{Filter-Then-Finetune} pipeline generates a large pool and applies filtering---combining automatic checks (e.g., LID consistency, perplexity, toxicity removal) with task-aware heuristics---before training the final model. \citet{Sravani2023EnhancingCTAE} demonstrate that training on a smaller, quality-filtered ``silver'' subset significantly outperforms using the full synthetic corpus.

\paragraph{Pseudo-parallel construction via monolingualization.}
\textbf{Parallel Corpus Construction} uses LLMs to convert code-mixed text into monolingual counterparts, creating pseudo-parallel pairs for supervised training \citep{Heredia2025ConditioningLTH}. This is particularly useful for objectives requiring aligned supervision (e.g., controlled switching or meaning preservation).

\paragraph{Constrained prompting for controlled, valid switching.}
For applications requiring strict controllability, \textbf{Linguistically Constrained Prompting} encodes theory-derived rules (e.g., allowable switch boundaries) directly into prompt. This ensures syntactically valid switching and pedagogical adherence while leveraging LLM fluency, as seen in frameworks like \textit{EZSwitch} \citep{Kuwanto2024LinguisticsTMG}.

\paragraph{Rule-based prompting for targeted pattern diversity.}
When the goal is to cover diverse mixing styles or to generate diagnostic patterns at scale, \textbf{Rule-Based Prompting} provides direct control (e.g., ``switch only nouns'', ``one switch per clause'', ``switch only inside NPs'').
This style is effective for generating balanced coverage over specific linguistic phenomena and for constructing challenge sets, as exemplified by approaches like \citet{Gupta2024CodeMixerYNL}.

\noindent\paragraph{Practitioner's Insight: The Data Generation Playbook}
\begin{itemize}
\item \textbf{Quality Beats Quantity:} Do not train on raw synthetic dumps. Follow the \textbf{``Filter-Then-Finetune''} protocol. A small, filtered ``silver'' dataset (screened for perplexity and switch density) outperforms a large, noisy one.
\item \textbf{For Unseen Language Pairs:} If you lack training data for a specific pair (e.g., Swahili-English), do not start from scratch. Use \textbf{GLOSS}-style adapters on a frozen multilingual backbone to transfer synthesis capabilities from high-resource pairs.
\item \textbf{Creating Supervised Signals:} If your downstream task requires parallel supervision (e.g., translation), use LLMs to ``monolingualize'' code-mixed text. This \textbf{Pseudo-Parallel Construction} creates the necessary aligned pairs that nature rarely provides.
\item \textbf{For Reliability and Diagnostics:} If generating data for or stress-testing, rely on \textbf{Constrained Prompting}. Use linguistic constraints (e.g., \textit{EZSwitch}) or rule-based substitution to guarantee validity of switch points, rather than relying on the model's ``vibe''.
\item \textbf{The Cost-Effective Baseline:} If compute is the bottleneck, revert to \textbf{Alignment-Guided Substitution}. It is cheap and fast, but apply POS/NER constraints to prevent the output from becoming grammatically broken.
\end{itemize}

\section{Training and Modeling Strategies}
\label{sec:Training}

Effectively handling code-mixed language requires strategies across the full model lifecycle, from \emph{foundational pre-training} to \emph{post-training adaptation}. Prior work shows that explicit exposure to code-mixing is essential: models trained with code-mixed objectives can outperform much larger models that lack such exposure \citep{zhang2023notyet}. We organize the literature into pre-training methods that inject code-mix awareness, and post-training methods that align and specialize the model for downstream use.

\subsection{Pre-training: Building Code-Mix-Aware LLMs}
Pre-training methods aim to bake code-mixing competence into the model so that language switching is handled as a first-class phenomenon rather than noise.

\paragraph{Objective design for code-mixing}
A common approach is to modify Masked Language Modeling (MLM) to emphasize the structured regions of code-mixed text. Boundary-aware objectives such as \textbf{SWITCHMLM} mask tokens around language switch points, while weakly supervised variants like \textbf{FREQMLM} use dictionary-based signals when gold labels are unavailable \citep{Das2023ImprovingPTAI}.
Other formulations add code-mix auxiliary tasks (e.g., switch-point prediction, bilingual generation) \citep{Baral2025CMLFormerADX} or cast learning as \textbf{denoising}, reconstructing monolingual text from synthetically code-mixed noise \citep{Iyer2023ExploringECAA}.

\paragraph{Architectures for preserving switch cues}
Architectural modifications can help preserve switch information that standard Transformers tend to smooth out. \textbf{RESBERT} adds residual shortcuts from intermediate layers to the MLM head to retain boundary information \citep{Das2023ImprovingPTAI}, while \textbf{CMLFormer} explicitly models matrix and embedded language interaction via a shared encoder and synchronized decoders \citep{Baral2025CMLFormerADX}.

\paragraph{Leveraging code-mix / switch in LLM pre-training to improve multilingual performance}
\textbf{CSCL} uses a staged curriculum (token-level $\rightarrow$ sentence-level $\rightarrow$ monolingual consolidation) to improve transfer and reduce forgetting \citep{Yoo2024CodeSwitchingCLK}.
In resource-constrained settings, \textbf{continued pre-training} on smaller models remains a strong baseline \citep{Goswami2023MixedDistilBERTCLZ}.

\noindent\paragraph{Practitioner's Insight: The Pre-Training Protocol}
\begin{itemize}
\item \textbf{Don't Mask Randomly:} Standard MLM is inefficient for code-mixing. Use \textbf{Boundary-Aware Masking} (e.g., SWITCHMLM) to focus model's gradient updates on the switch-points, where grammatical complexity lies.
\item \textbf{Preserve the Signal:} Deep transformers tend to ``smooth out'' the distinction between languages. If training smaller BERT-style models, add \textbf{Residual Connections} (RESBERT) to keep language-id signals available at the output layer.
\item \textbf{Order Matters (Curriculum):} Do not dump all code-mixed data at once. Follow a \textbf{Staged Curriculum} (CSCL): start with token-level mixing to learn vocabulary alignment, then move to sentence-level switching to learn syntax, and finish with monolingual consolidation to prevent catastrophic forgetting and benefit the overall multilingual capabilities of the system.
\end{itemize}

\subsection{Post-training: Specializing, Aligning, and Controlling Behavior}
Post-training methods start from a general base model and adapt it to downstream tasks, deployment objectives, and desired code-mixing behavior.

\paragraph{Supervised and instruction tuning}
\textbf{Supervised fine-tuning (SFT)} on labeled or pseudo-parallel data is the standard approach for task specialization, including conditional generation such as monolingual $\rightarrow$ code-mixed translation \citep{Heredia2025ConditioningLTH}.
For parameter-efficient specialization, PRO-CS utilizes instance-based prompt composition to combine task and language knowledge, achieving performance on par with full fine-tuning using only 0.18\% of total parameters \citep{bansal-etal-2022-pro}. \textbf{Instruction tuning} improves controllability; \textbf{COMMIT} keeps instruction templates in English while code-mixing content fields, preserving instruction clarity while exposing model to mixed inputs \citep{lee-etal-2024-commit}.

\paragraph{Preference alignment, personalization, and cross-lingual transfer}
Beyond SFT, \textbf{RLAIF} aligns outputs with human preferences via reward modeling and RL, optimizing directly for perceived quality \citep{zhang2025improvingcodemixed}.
Fine-tuning can also target stylistic control: \textbf{PARADOX} conditions generation on a learned user persona to match individual mixing styles \citep{Sengupta2023PersonaawareGMAB}.
For transferring domain competence across languages, \textbf{MIGRATE} leverages code-switching data to adapt domain-specific English models efficiently \citep{hong-etal-2025-migrate}, and synthetic code-switched tuning can improve low-resource performance without degrading high-resource languages \citep{Nagar2025BreakingLanguageBarriers}.

\paragraph{Representations and structure for downstream robustness}
Post-training also includes techniques that shape representations for reliable downstream use. \textbf{Contrastive learning} objectives such as \textbf{ConCSE} align code-mixed sentences with monolingual equivalents to produce semantically stable embeddings \citep{Jeon2024ConCSEUCAH}.
For retrieval settings, \textbf{ContrastiveMix} decouples language alignment from relevance training to preserve ranking signals in IR/RAG \citep{do-etal-2024-contrastivemix}.
For structure-sensitive tasks, explicit syntax can be incorporated via \textbf{code-mixed UD forests}, which merge parses from both languages into a unified graph to reduce single-parser transfer bias in relation extraction \citep{Fei2023ConstructingCUBP}.

\paragraph{Practitioner's Insight: The Adaptation Playbook}
\begin{itemize}
\item \textbf{Instruction Tuning Design:} Do not code-mix the instructions themselves. Follow the \textbf{COMMIT} strategy: keep the task template (instruction) in a high-resource language (like English) and only code-mix the input/output content. This prevents the model from misunderstanding the task definition.
\item \textbf{Efficiency via Soft Prompts:} Use \textbf{Prompt Composition} (PRO-CS) to fuse language and task embeddings. This achieves parity with full fine-tuning using less than 1\% of the parameters, essential for deploying multiple language pairs.
\item \textbf{Quality requires Alignment, not just SFT:} SFT teaches the model \textbf{how} to mix, but not \textbf{what is good}. To fix ``unnatural'' mixing, you must use \textbf{RLAIF} (Reinforcement Learning with AI Feedback) or reward modeling to align the model with preference scores rather than just next-token prediction.
\item \textbf{Personalization:} Use \textbf{Persona-Aware Generation} (PARADOX) by conditioning the generation on a user-specific embedding or style profile.
\item \textbf{Fixing Retrieval (RAG):} Standard multilingual embeddings often fail to match code-mixed queries to monolingual documents. Use \textbf{Contrastive Decoupling} (ContrastiveMix) to separate ``language alignment'' loss from ``relevance'' loss, ensuring the ranker doesn't get confused by language shifts.
\end{itemize}

\noindent For a detailed discussion of embedding methods for code-mixed language---including contrastive alignment (ConCSE, ContrastiveMix), switch-aware architectures (CMLFormer, RESBERT), and structure-informed embeddings (UD Forests)---see Appendix~\ref{app:Embeddings}.

\section{Evaluation: From Overlap Metrics to Capability Probes}
\label{sec:Evaluation}

Evaluation in the LLM era has shifted from surface-level overlap (does the output match a reference?) to capability assessment (does the model preserve meaning, follow mixing constraints, and remain useful?). This shift is driven by a persistent \textbf{metric gap}: conventional n-gram metrics (BLEU, ROUGE) are insensitive to code-mixing pathologies such as \emph{language reversion} (collapsing to English) or \emph{uncontrolled switching} \citep{zhang2023notyet}. Current best practices therefore decouple the measurement of \emph{structural mixing} from \emph{semantic quality}, organizing benchmarks by the specific reasoning capabilities they stress-test.

\subsection{Measuring Quality: A Hybrid Assessment Strategy}
Evaluating code-mixed text requires more than standard overlap metrics, since there are many valid bilingual realizations and metrics like BLEU can over-reward monolingual copying. A practical evaluation setup therefore combines two complementary views: \emph{structural instrumentation} to verify that mixing is actually happening, and \emph{semantic judgment} to ensure the output remains faithful, natural, and useful.

\paragraph{Structural Indices as Instrumentation.}
Structural metrics help diagnose failure modes that overlap metrics miss, especially when models silently revert to a single language. The \textbf{Code Mixing Index (CMI)} \citep{aguilar2020lince} measures how much mixing occurs, making it a useful first check for low-effort outputs or language collapse. However, volume alone is insufficient because unnatural mixing can still score well. The \textbf{I-Index} (Switching Ratio) captures switching frequency \citep{srivastava-singh-2021-challenges}, helping distinguish coherent span-level switching from pathological token-level flickering \citep{Deng2025SparseASI}. Complementary measures such as the \textbf{M-Index} further help separate meaningful bilingual mixing from shallow lexical borrowing.

\paragraph{Human-Centric and AI Judges.}
For semantic faithfulness and naturalness, reference-free evaluation is often more reliable than n-gram matching. \textbf{Likert-scale acceptability} remains the most direct way to assess whether generations align with bilingual community norms \citep{Kodali2024FromHJAP}, while \textbf{pairwise preference} comparisons tend to be more stable when selecting between competing prompting or training strategies \citep{Heredia2025ConditioningLTH}. To scale human insights, \textbf{LLM-as-a-Judge} protocols are increasingly used \citep{zhang2025improvingcodemixed}, but they must be calibrated against bilingual human ratings to avoid bias toward fluent monolingual outputs.

\noindent\paragraph{Practitioner's Insight: The Metric Selection Playbook}
\begin{itemize}
\item \textbf{Treat structure and meaning as separate checks:} Use overlap metrics for rough fluency, but always pair them with structural indices to confirm real code-mixing rather than monolingual copying.
\item \textbf{Prefer human-aligned judgments for generation:} For open-ended generation, \textbf{pairwise preference} and \textbf{Likert scoring} reflect perceived naturalness and faithfulness more reliably than exact match metrics.
\item \textbf{Calibrate LLM judges before trusting them:} Validate judge scores on a small bilingual human-labeled subset, and track judge disagreement with humans as a first-class quality signal.
\end{itemize}

\noindent A detailed comparison of evaluation method families is provided in Appendix~\ref{app:EvalTable}.

\subsection{Benchmarking: Selecting the Right Testbed}

Early datasets focused on shallow tagging; modern evaluation should probe reasoning, generalization, and knowledge activation.

\paragraph{Foundational Syntax and Understanding.}
For baseline linguistic competence, \textbf{LinCE} \citep{aguilar2020lince} and \textbf{GLUECoS} \citep{khanuja-etal-2020-gluecos} provide standardized splits for Language ID, POS tagging, and NLI. Recent large-scale resources like \textbf{COMI-LINGUA} \citep{sheth2025comilinguaexpertannotatedlargescale} extend this to multi-task settings. However, high accuracy on these benchmarks can be deceptive; models often take ``monolingual shortcuts'' by translating internally or ignoring the mixed context entirely.

\paragraph{Reasoning, Knowledge, and Generation.}
To quantify ``Code-Mixing Gap'' in complex tasks, suites like \textbf{CodeMixEval} \citep{Yang2025EvaluatingCIE} and \textbf{Lost in the Mix} \citep{Mohamed2025LostITD} isolate specific reasoning degradation caused by switching languages. In retrieval, benchmarks like \textbf{ENKOQA} test whether code-mixed queries ``activate'' cultural knowledge better than English equivalents \citep{Kim2024CanCTO}. For summarization, \textbf{CS-Sum} reveals high automatic scores mask subtle, meaning-altering errors, necessitating fine-grained error taxonomies \citep{Suresh2025CSSum}.

\paragraph{Diagnostic Probes.}
Advanced evaluation targets specific linguistic failures. \textbf{Minimal-pair evaluation} determines if models respect grammatical constraints (e.g., prohibiting switches within bound morphemes) \citep{Sterner2025MinimalPEQ}. Synthetic \textbf{NLI probes} measure the stability of logical alignment across language boundaries \citep{Khandelwal2025AlignNLI}, while other diagnostics target low-resource transfer efficacy \citep{Parra2023Basquenglish}.

\noindent\paragraph{Practitioner's Insight: Benchmarking Playbook}
\begin{itemize}
\item \textbf{Build a three tier eval suite:} (1) competence checks (LinCE, GLUECoS), (2) end task gap measurement (CodeMixEval, Lost in the Mix, ENKOQA, CS-Sum), and (3) diagnostics (Minimal Pairs, NLI probes). This prevents overfitting to shallow benchmarks.
\item \textbf{Report the Code Mixing Gap explicitly:} Always compare performance on code-mixed inputs versus matched monolingual inputs (same intent, same entities). The gap is often more informative than absolute scores.
\item \textbf{Stratify by mixing difficulty:} Slice results by switch rate, script type (native vs romanized), and switch location (token boundary vs intra word). Aggregated numbers hide systematic failures.
\end{itemize}

\section{Safety, Bias, and Red-Teaming in Code-Mixed Contexts}
\label{sec:Safety}

The fluid complexity of code-mixing presents unique challenges to model safety, often rendering guardrails trained on monolingual, high-resource data ineffective. This section examines how code-switching (CSW) serves as an adversarial attack vector and highlights the systemic biases inherent in current data practices.

\subsection{Code-Switching as an Attack Vector}
Research consistently demonstrates that intra-sentence code-switching is a potent method for circumventing safety filters. The \textbf{Code-Switching Red-Teaming (CSRT)} framework reveals that LLMs frequently comply with harmful requests in code-mixed formats that they would refuse in English, with Attack Success Rate (ASR) increasing as more languages---particularly low-resource ones---are interleaved \citep{Yoo2024CodeSwitchingRLR}.
This vulnerability stems from \textbf{lexical dilution} weakening keyword defenses. Attack efficacy is amplified by automated toxicity generators \citep{WadgaonkarRatha2025AdversarialToxicity} and phonetic perturbation (e.g., ``Haet Bhasha''), which masks toxic intent via noisy romanization to bypass filters \citep{aswal2025haetbhashaaurdiskrimineshun}.
Mitigating them requires moving beyond English-only safety alignment to implement dynamic, CSW-aware guardrails and explicitly including code-mixed adversarial examples in safety training.

\subsection{Bias, Representativeness, and Future Scopes}
Beyond security, ethical concerns arise from the data pipelines themselves. \citet{Doruz2023RepresentativenessAA} argue that the lack of diversity in data collection and annotation teams introduces systematic blind spots regarding dialects and socio-demographics. To mitigate downstream harms, they advocate for transparent documentation and checklists to ensure representativeness across geographical and social registers.

\section{Future Research: Grand Challenges}
\label{sec:Future}

We highlight four grand challenges that shape the next stage of CSW research.

\paragraph{Challenge 1: Breaking the English-Centric Data Barrier.}
CSW pipelines remain skewed toward English-paired mixing \citep{Doruz2023ASOU, Sterner2024MultilingualIOAQ}. Scaling beyond coverage into \emph{representativeness}---reflecting dialectal variation, socio-demographic diversity, and realistic usage---is essential \citep{Doruz2023RepresentativenessAA}.

\paragraph{Challenge 2: Reforming the Evaluation Paradigm.}
Current metrics often reward language reversion and miss meaning drift under switching \citep{Heredia2025ConditioningLTH}. Progress requires reproducible protocols that jointly assess structural validity and semantic faithfulness across mixing regimes, with LLM judges calibrated against bilingual human acceptability \citep{Kuwanto2024LinguisticsTMG, Kodali2024FromHJAP}.

\paragraph{Challenge 3: RL, Reasoning, and Long-Context Modeling.}
Reinforcement learning remains underexplored for CSW-specific objectives (faithfulness, naturalness, controllable switching). Long-context CSW is poorly understood, especially for multi-turn consistency and grounded reasoning. Mixed-language reasoning traces arise naturally in LLMs and may improve reasoning; future RL objectives could leverage these traces rather than suppressing them.

\paragraph{Challenge 4: Dynamic Safety Under Mixed Inputs.}
Code-mixing bypasses English-centric safeguards through lexical dilution \citep{aswal2025haetbhashaaurdiskrimineshun}. Developing mix-aware guardrails requires adversarial evaluation reflecting realistic attacks and mechanistic interpretability to identify where mixed representations cause safety failures \citep{Nie2025MechanisticUAAO, Baral2025CMLFormerADX}.

\section{Conclusion}

This survey provides a comprehensive overview of CSW challenges in the LLM era. Our analysis shows that effective CSW modeling demands explicit interventions: CSW-aware pre-training, post-training adaptations, and mechanistic control. Standard automatic metrics are unreliable; the field must shift toward human-centered evaluation.
We propose a roadmap requiring community efforts to expand data diversity, reform evaluation, and develop CSW-aware safety alignments. 

\section*{Limitations}
This survey focuses on the LLM lifecycle for code-mixing and code-switching and therefore does not exhaustively cover pre-LLM statistical and neural approaches, speech-only settings, or non-textual modalities. Our coverage of language pairs reflects the English-centric bias of the literature we review, and we highlight this as an open problem rather than claiming full typological coverage. Finally, as a survey, we do not introduce new empirical results; conclusions about relative effectiveness of methods inherit the limitations of the underlying studies.

\bibliography{custom}

\clearpage

\appendix

\section{Landscape of Tasks and Applications}
\label{app:Tasks}

Code-mixing is not a single task but a pervasive linguistic condition that reshapes requirements across the NLP stack. Because mixed-language inputs interleave vocabularies, scripts, and grammatical systems within the same utterance, systems must support foundational analysis tasks such as token and span level language identification, part of speech tagging, and named entity recognition, along with deeper structural processing such as dependency parsing and syntax aware relation extraction.

At the semantic level, code-mixing affects knowledge intensive applications including search and retrieval, semantic similarity, and question answering, where models must remain robust to query document mismatch across scripts and languages. It also impacts generation oriented settings such as machine translation to and from code mixed text, controlled generation with natural switch patterns, and interactive assistants for dialogue and summarization, as well as education focused tasks like learner friendly grammatical error correction that preserves intended switches. Across these families, code-mixing amplifies cross cutting concerns around personalization, reliability, and safety, including robustness and guardrail bypass under mixed language prompts.

\section{Evaluation Method Comparison}
\label{app:EvalTable}

Table~\ref{tab:eval_methods} summarizes the primary evaluation method families for code-mixed LLMs, contrasting their signals, strengths, and limitations. Reliable assessment typically requires combining structural instrumentation with semantic and acceptability signals.

\begin{table*}[t]
\centering
\footnotesize
\setlength{\tabcolsep}{5pt}
\renewcommand{\arraystretch}{1.1}
\caption{Comparison of evaluation methods for code-mixed LLMs. Each method emphasizes a different construct; reliable assessment typically requires combining structural instrumentation with semantic and acceptability signals.}
\label{tab:eval_methods}
\begin{tabular}{p{3.2cm} p{4.0cm} p{6.8cm}}
\toprule
\textbf{Method Family} & \textbf{Primary Signal} & \textbf{Strengths \& Limitations} \\
\midrule

\textbf{Overlap metrics} \textit{(BLEU, ROUGE)} 
& Reference n-gram overlap 
& \textbf{+} Exact-match, translation-like tasks \textbf{--} Rewards monolingual copying; weak switch sensitivity \\

\textbf{Structural indices} \textit{(CMI, M-Index, I-Index)} 
& Language IDs \& token statistics 
& \textbf{+} Detects collapse, imbalance, over/under-switching \textbf{--} No semantic or naturalness assessment \\

\textbf{Human acceptability} \textit{(Likert)} 
& Bilingual naturalness \& grammaticality 
& \textbf{+} Identifies ungrammatical/odd switching \textbf{--} Expensive; inter-rater \& community variation \\

\textbf{Pairwise preference} 
& Relative output quality 
& \textbf{+} Robust model/prompt selection \textbf{--} No absolute quality scores \\

\textbf{LLM-as-a-Judge} 
& Rubric-based LLM scoring 
& \textbf{+} Fast regression testing \& semantic evaluation \textbf{--} Judge bias; needs calibration \\

\textbf{Targeted semantic} \textit{(e.g., PIER)} 
& Word/slot/entity-level correctness 
& \textbf{+} Switch-local semantic errors \textbf{--} Needs predefined points; misses global coherence \\

\bottomrule
\end{tabular}
\end{table*}

\section{Embeddings for Code-Mixed Language}
\label{app:Embeddings}

A central challenge in modeling code-mixed text is the creation of embeddings that maintain semantic consistency when languages and scripts are intermingled. The goal is to ensure that a sentence and its code-switched counterpart are mapped to nearby points in the vector space, a property not guaranteed by standard multilingual models. Research in this area has moved beyond simply applying pre-trained models to developing specialized objectives and architectures designed to produce robust, switch-aware code-mix embeddings.

\subsection{Contrastive Learning for Cross-Lingual Alignment}
A primary strategy for creating robust code-mix embeddings is contrastive learning, which explicitly trains a model to align the representations of monolingual sentences with their code-switched equivalents.

The seminal work in this area is \textbf{ConCSE}, a framework that extends SimCSE by introducing cross-contrastive and triplet loss objectives \citep{Jeon2024ConCSEUCAH}. During training, the model is tasked with pulling the embedding of a monolingual sentence closer to its code-mixed version while pushing it away from other negative examples. This process directly optimizes the embedding space for consistency across monolingual and code-switched forms, resulting in sentence embeddings that are significantly more robust for downstream tasks like semantic textual similarity.

For the specific domain of Information Retrieval (IR), \citet{do-etal-2024-contrastivemix} identified a key problem: naively training on code-mixed data can improve language alignment at the expense of relevance matching. Their proposed method, \textbf{ContrastiveMix}, disentangles these two forces with a dual-loss architecture. It retains the standard IR contrastive loss for query-passage relevance while adding a separate, query-side loss to align the embeddings of English and code-mixed query variations. This approach successfully creates embeddings that improve cross-lingual retrieval without degrading the core relevance-matching capability.

\subsection{Architectures and Pre-training for Switch-Aware Embeddings}
Another major line of work focuses on modifying pre-training objectives and model architectures to build switch-awareness directly into the encoder. Instead of aligning representations post-hoc, these methods produce embeddings that are inherently sensitive to the structure of code-mixing.

\textbf{CMLFormer} introduces a novel pre-training setup with a shared encoder and two synchronized decoders, augmented with switch-specific tasks like predicting the location of language switches \citep{Baral2025CMLFormerADX}. When detached for downstream tasks, the shared encoder produces token and sentence embeddings that are explicitly informed by the code-mixing seen during pre-training, leading to improved performance on tasks like hate speech detection.

\textbf{RESBERT} enhances the standard BERT architecture with two key modifications: a residual connection from an intermediate layer into the final MLM head and an auxiliary token-level Language ID loss \citep{Das2023ImprovingPTAI}. Probing experiments show that the resulting hidden states better encode switch-boundary information, creating an embedding space where code-mixed features are more localized and separable. A simpler but effective approach is \textbf{continued pre-training}, where a model like DistilBERT is further trained on synthetic code-mixed text. This was shown by \citet{Goswami2023MixedDistilBERTCLZ} to yield more stable embeddings under the noisy and transliterated conditions common in South Asian code-mixed text.

\subsection{Embeddings from Explicit Linguistic Structures}
A more specialized but powerful approach involves creating embeddings that are informed by explicit linguistic structure from multiple languages simultaneously. The most prominent example is the use of \textbf{Code-Mixed Universal Dependency (UD) Forests} for cross-lingual relation extraction \citep{Fei2023ConstructingCUBP}. This method first generates UD parse trees for a sentence in both its source and target languages. These two trees are then merged into a single graph structure, or ``forest.'' A Graph Attention Network then encodes this forest, creating node embeddings that combine the semantic information from a multilingual language model with the rich, cross-lingual topological information from the merged parse trees. The resulting embeddings are syntactically aware across both languages and have been shown to significantly improve performance by mitigating transfer bias.

\end{document}